\documentclass[conference]{IEEEtran}
\IEEEoverridecommandlockouts
\usepackage{cite}
\usepackage{url}
\usepackage{amsmath,amssymb,amsfonts}
\usepackage{algorithmic}
\usepackage{graphicx}
\usepackage{textcomp}
\usepackage{xcolor}
\usepackage{subcaption}
\usepackage{multirow}
\usepackage{enumitem} 
\usepackage{tikz}
\usepackage{makecell}
\def\BibTeX{{\rm B\kern-.05em{\sc i\kern-.025em b}\kern-.08em
    T\kern-.1667em\lower.7ex\hbox{E}\kern-.125emX}}
    
\begin{document}

\title{Fragile Watermarking for Image Certification Using Deep Steganographic Embedding
}

\author{\IEEEauthorblockN{Davide Ghiani$^1$, Jefferson David Rodriguez Chivata$^1$, Stefano Lilliu$^1$, Simone Maurizio La Cava$^1$,\\ Marco Micheletto$^1$, Giulia Orr\'u$^1$, Federico Lama$^2$, Gian Luca Marcialis$^1$}
\IEEEauthorblockA{$^1$University of Cagliari, Piazza d'Armi I - 09123 Cagliari (Italy), e-mail: \\
\{davide.ghiani,jeffersond.rodriguez,simonem.lac,marco.micheletto,giulia.orru,marcialis\}@unica.it}
\IEEEauthorblockA{$^2$Dedem S.p.A., Via Cancelleria 59 - 00072 Ariccia (Italy), e-mail: federico.lama@dedem.it
}}

\newcommand\copyrighttext{%
\footnotesize \textcopyright 2025 IEEE. Personal use of this material is permitted.
Permission from IEEE must be obtained for all other uses, in any current or future
media, including reprinting/republishing this material for advertising or promotional
purposes, creating new collective works, for resale or redistribution to servers or
lists, or reuse of any copyrighted component of this work in other works.}
\newcommand\copyrightnotice{%
\begin{tikzpicture}[remember picture,overlay]
\node[anchor=south,yshift=10pt] at (current page.south) {\fbox{\parbox{\dimexpr\textwidth-\fboxsep-\fboxrule\relax}{\copyrighttext}}};
\end{tikzpicture}%
}

\maketitle
\copyrightnotice

\begin{abstract}
Modern identity verification systems increasingly rely on facial images embedded in biometric documents such as electronic passports. To ensure global interoperability and security, these images must comply with strict standards defined by the International Civil Aviation Organization (ICAO), which specify acquisition, quality, and format requirements. However, once issued, these images may undergo unintentional degradations (e.g., compression, resizing) or malicious manipulations (e.g., morphing) and deceive facial recognition systems.
In this study, we explore fragile watermarking, based on deep steganographic embedding as a proactive mechanism to certify the authenticity of ICAO-compliant facial images. By embedding a hidden image within the official photo at the time of issuance, we establish an integrity marker that becomes sensitive to any post-issuance modification. We assess how a range of image manipulations affects the recovered hidden image and show that degradation artifacts can serve as robust forensic cues. Furthermore, we propose a classification framework that analyzes the revealed content to detect and categorize the type of manipulation applied. Our experiments demonstrate high detection accuracy, including cross-method scenarios with multiple deep steganography-based models. These findings support the viability of fragile watermarking via steganographic embedding as a valuable tool for biometric document integrity verification.
\end{abstract}


\section{Introduction}
Nowadays, facial recognition (FR) technology plays a crucial role in identity verification, serving as a fundamental component in border control, secure identity management, and forensic applications \cite{la20233d}. To enhance security and streamline passenger processing, many countries have adopted electronic passports (ePass), which store biometric data to enable accurate and automated identity verification at border checkpoints.
According to the International Civil Aviation Organization (ICAO) guidelines, facial images stored in machine readable travel documents, such as ePass, must comply with strict biometric standards to facilitate reliable authentication \cite{icao20159303}. 

However, multiple factors can affect the reliability of these systems, including compression artifacts, noise, or other distortions introduced during acquisition, transmission, or storage \cite{10.1145/3507901}. In addition to these non-malicious alterations, the increasing sophistication of image manipulation techniques has introduced new security vulnerabilities that can compromise the integrity of identity verification \cite{Tolosana2022}.

One of the most pressing threats is the morphing attack, which leverages image synthesis techniques to blend facial features of multiple subjects, generating realistic composite images that can be falsely considered to belong to different individuals \cite{scherhag2019face}. This vulnerability is particularly critical in border security, where identity verification is based on comparing a live subject with the ePass photograph. If a morphed image is successfully enrolled in an ePass, both contributing individuals can authenticate using the same document, bypassing security checks \cite{robertson2017fraudulent, ferrara2014magic}.
To mitigate the risks associated with morphing attacks, Morphing Attack Detection (MAD) techniques have been developed to differentiate between genuine and manipulated facial images \cite{ngan2020face}. Despite significant progress, existing MAD methods face challenges in terms of generalization across novel morphing techniques, and adaptability to real-world conditions of identity recognition scenarios \cite{venkatesh2021face, panzino2024evaluating}.

To address these limitations, researchers have investigated proactive mechanisms that embed verification signals within the image itself at acquisition time, ensuring integrity throughout the document lifecycle. A promising direction involves active authentication mechanisms that introduce integrity markers directly within an image to facilitate the detection of manipulation. Among such techniques, digital watermarking has been widely adopted in multimedia security \cite{song2024survey}; however, watermarking is designed primarily for copyright protection and may lack the adaptability required for ePass applications, where facial images must remain unchanged after issuance.
In parallel, deep learning-based steganography has recently enabled the embedding of large amounts of data into cover images with minimal visual distortion. 
Although steganography is not designed for integrity verification, its capacity and perceptual quality open opportunities for alternative use cases.
In this work, we hypothesize that steganographic models can be repurposed to implement active integrity verification mechanisms for facial images. Specifically, we propose a fragile watermarking framework based on deep steganographic embedding, in which any manipulation of the host image degrades the embedded content, allowing post-hoc integrity verification through reconstruction of a known marker.

The concealed image can only be retrieved through a dedicated decoding process, which generates a revealed image. Since the hidden data is embedded within the cover image, any modification applied to such an image will inevitably affect the revealed image, therefore introducing artifacts that could potentially provide a forensic indicator of tampering \cite{huang2024robust}. 

Despite its long history in multimedia applications, the use of steganography-inspired embedding methods for forensic integrity verification in ICAO-compliant biometric documents has, to the best of our knowledge, yet to be explored. This represents a novel direction in the fight against digital attacks such as morphing: we hypothesize that tampering leaves subtle but detectable fingerprints in the revealed marker.

In this regard, this study proposes a twofold contribution: (i) assessing the feasibility of fragile watermarking via deep steganographic embedding as a mechanism to verify the integrity of ICAO-compliant facial images, ensuring that any post-issuance modifications, whether intentional or accidental, can be detected; (ii) developing a classification model capable of distinguishing between different types of alterations, such as morphing, compression artifacts, and noise addition, based on how these transformations affect the retrieved marker.

The aim is not to develop a novel steganographic method, but rather to investigate whether standard steganographic models, when repurposed as fragile watermarking tools, inherently offer resilience properties applicable to biometric image security. By examining how various manipulations impact the fidelity of the revealed content, we explore the potential of this approach for unauthorized modification detection.

The rest of this paper is organized as follows. Section \ref{sec:related} reviews the current literature on steganography and watermarking for digital images. Section \ref{sec:approach} describes the proposed approach. Section \ref{sec:protocol} reports the experimental protocol employed to conduct our evaluation, while Section \ref{sec:results} reports the obtained results. Finally, conclusions are drawn in Section \ref{sec:conclusions}.

\section{Related Work}\label{sec:related}

Steganography and watermarking are both data hiding frameworks that can be classified according to their tolerance to image modifications \cite{wang2023data}. While fragile approaches are highly sensitive and signal any alteration in the host image, robust methods aim to resist various manipulations, including synthetic content generation such as deepfakes \cite{zhao2023proactive}. 
In the context of ICAO-compliant facial verification, fragile mechanisms are preferable, as their universal sensitivity enables the detection of any post-acquisition modification. This aligns with the proactive nature of ICAO security requirements, where even minor unauthorized alterations must be detected to preserve the integrity of biometric documents.
Following this, it is important to distinguish how the information is embedded and interpreted in the frameworks.  Digital watermarking typically relies on compact bitstrings embedded at predefined locations or patterns. These are effective in robust scenarios like copyright protection \cite{wang2024invisible} or tamper detection \cite{zhang2024editguard}, but often lack interpretability when the watermarked image is degraded or tampered with.
Conversely, modern steganographic models enable the embedding of richer information, such as entire images, directly into the visual structure of the host. Originally developed for covert messaging, these techniques can be repurposed for fragile watermarking. In particular, any modification to the host image corrupts the embedded payload, and this degradation can be visually observed in the recovered content, providing actionable forensic indicators of tampering. This repurposing is made possible by recent advances in deep learning, which have significantly improved the imperceptibility and recoverability of hidden information. Architectures based on convolutional neural networks (CNNs) \cite{zhu2018hidden}, generative adversarial networks (GANs) \cite{zhang2019steganogan}, and autoencoders \cite{ke2024stegformer} have demonstrated high-capacity and visually stable embedding and extraction, making them suitable candidates for integrity verification in fragile settings.

\section{Proposed Approach}\label{sec:approach}

\begin{figure*}
    \centering
    \includegraphics[width=0.75\textwidth]{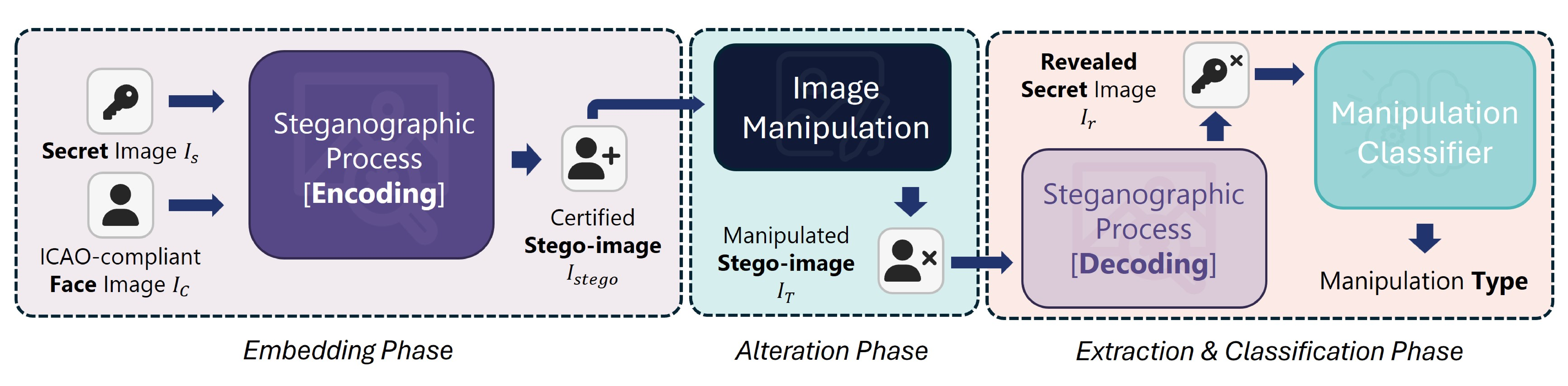}
    \caption{Overview of the proposed methodology, structured into three main phases: embedding, alteration and classification.}
    \label{fig:proposed_method}
\end{figure*}

The proposed approach introduces a fragile watermarking mechanism, implemented through deep steganographic embedding, to certify the authenticity of ICAO-compliant facial images, ensuring long-term verifiability of photographs used in official identity systems by embedding a hidden integrity marker at the time of issuance. If the stego-image remains unaltered, the embedded content can be retrieved without distortion. Conversely, any modification to the stego-image, whether due to standard image processing or deliberate tampering, inevitably affects the extracted content.
Our key hypothesis is that different manipulations introduce systematic and detectable artifacts in the recovered marker, which can signal both the presence and nature of the transformation. If these degradation patterns are consistent, they can serve as a forensic cue for tamper detection and manipulation classification. To evaluate this hypothesis, we define a three-stage methodology (Figure \ref{fig:proposed_method}):
\begin{enumerate}[leftmargin=*]
    \item A steganographic process $E$ is applied to the ICAO-compliant facial image (cover image) $I_{C}$ to embed a secret marker image $I_{S}$ within it, producing a certified watermarked image $I_{stego}$:
    \begin{equation}
    \small
        I_{stego} = E \left ( I_{C}, I_{S} \right )
    \end{equation}
The hidden image $I_S$, imperceptibly hidden within the cover image, acts as a fragile integrity marker, ensuring that any future modification to the stego-image affects the embedded content.
    \item A set of controlled transformations $T$ are applied to $I_{stego}$ to simulate real-world manipulations:
    \begin{equation}
    \small
        I_t = T(I_{stego})
    \end{equation}
        The considered manipulations—resizing, compression, noise, blur, sharpening, and morphing—reflect both common post-processing operations and intentional biometric attacks. These transformations simulate real-world conditions where an image might be altered after issuance, allowing us to assess whether the hidden integrity marker can act as a forensic signal.

    \item A decoder $D$ is used to extract the revealed image $I_r$ from the potentially modified image $I_t$:
    \begin{equation}
    \small
        I_r = D(I_t)
    \end{equation}
If the stego-image remains unchanged (i.e., no transformation artifacts), the revealed image retains its expected structure. However, when modifications occur, artifacts emerge, reflecting the type and severity of the applied transformation. To systematically analyze these distortions, we introduce a classification model capable of distinguishing between different manipulation patterns. \end{enumerate}

The proposed system assumes that the integrity marker is present and matches the expected reference at verification time. In scenarios where the marker is missing, mismatched, or unrecoverable, the classification model may yield unreliable outputs. Handling such cases requires a separate detection stage, which is not addressed in the current work.

The next section presents the complete experimental pipeline, detailing the embedding architecture, transformation setup, classifier design, and evaluation criteria.

\section{Experimental protocol}\label{sec:protocol}

\begin{figure}
\centering
    \begin{subfigure}{0.09\textwidth}
        \includegraphics[width=\linewidth]{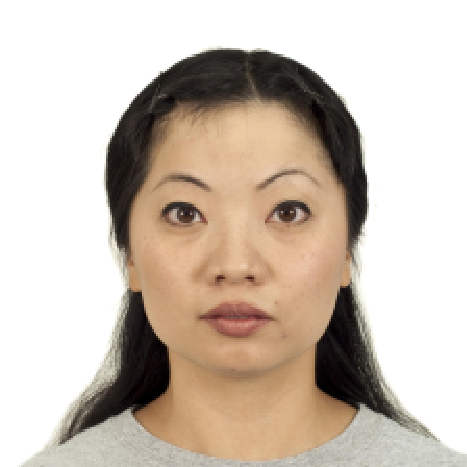}
        \caption{}
    \end{subfigure}
    \begin{subfigure}{0.09\textwidth}
        \includegraphics[width=\linewidth]{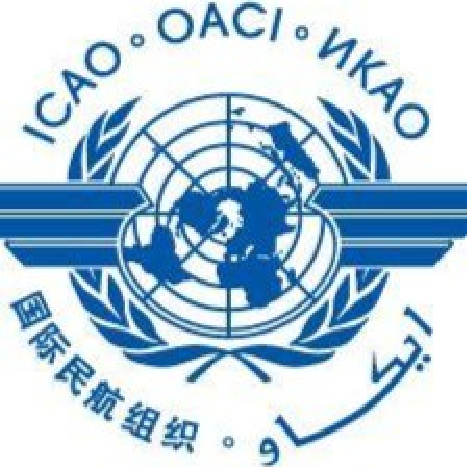}
        \caption{}
    \end{subfigure}\\
        \begin{subfigure}{0.09\textwidth}
        \includegraphics[width=\linewidth]{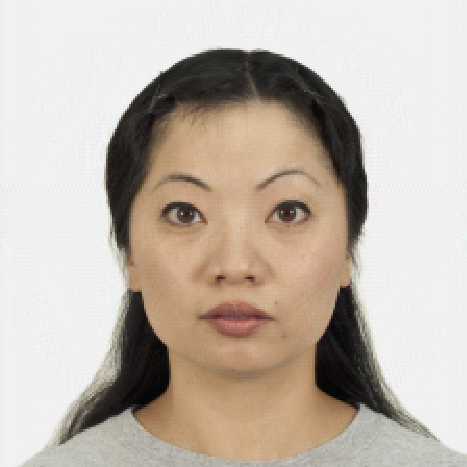}
        \caption{}
    \end{subfigure}
    \begin{subfigure}{0.09\textwidth}
        \includegraphics[width=\linewidth]{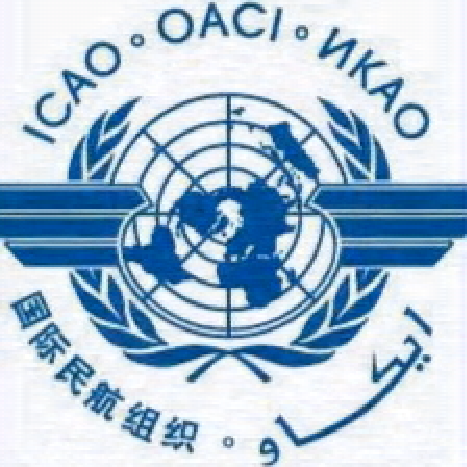}
        \caption{}
    \end{subfigure}
    \begin{subfigure}{0.09\textwidth}
        \includegraphics[width=\linewidth]{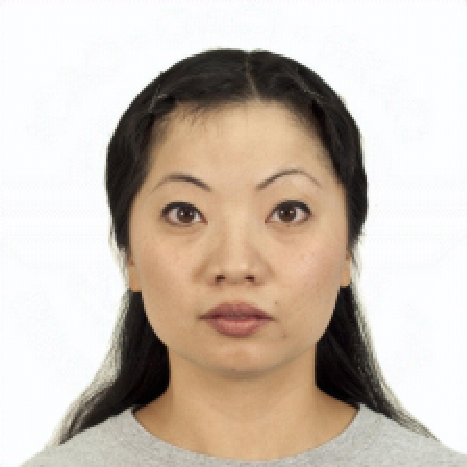}
        \caption{}
    \end{subfigure}
    \begin{subfigure}{0.09\textwidth}
        \includegraphics[width=\linewidth]{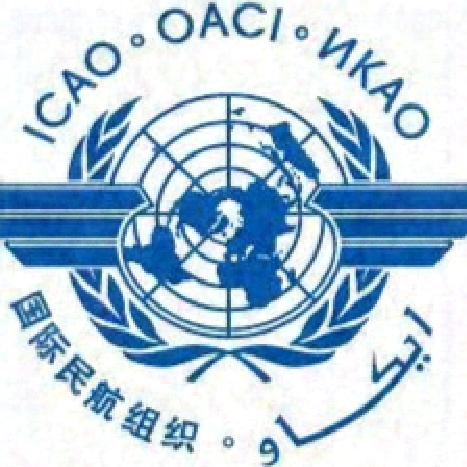}
        \caption{}
    \end{subfigure}
    \caption{Overview of the steganographic certification process and its impact on image quality: a) original input image;  b) original secret image to be embedded; c) stego image generated using Steguz; d) secret image recovered from the Steguz stego image;  e) stego image generated using Stegformer; f) secret image recovered from the Stegformer stego image. }
    \label{fig:certification_examples}
\end{figure}

\subsection{Dataset}

The goal is to assess the feasibility of our fragile watermarking approach on facial images compliant with ICAO guidelines. Accordingly, we selected the Chicago Face Database (CFD) \cite{ma2015chicago, ma2021chicago, lakshmi2021india}, providing high-quality facial images of 827 men and women of varying ethnicity between the ages of 17-65. Specifically, it includes a single ICAO-compliant image per subject \cite{seibold2020accurate}, with a $2444 \times 1718$ resolution.

To prepare the data for embedding, each image was cropped to a $1718\times1718$ square format, removing peripheral background while retaining the facial region. The cropped images were then resized to $224\times224$ pixels to comply with the input requirements of the steganography-based models used for analysis.
As the integrity marker, we selected the ICAO logo, which was resized to $224 \times 224$ and embedded into each subject’s facial image during the certification phase (Figure \ref{fig:certification_examples}).

\subsection{Steganography Models}

\label{subsec:stega}
As discussed in Section \ref{sec:related}, recent deep learning-based steganography methods offer high-capacity and high-fidelity embedding mechanisms. Although originally developed for covert communication, these models can be repurposed for fragile watermarking, enabling integrity verification through the degradation of a hidden marker.
To investigate the generality of our approach and support cross-model comparisons, we selected two representative state-of-the-art methods. The first, \textit{Stegformer} \cite{ke2024stegformer}, is a transformer-based autoencoder architecture designed for dense image-to-image embedding. The second, \textit{SteGuz} \cite{khalifa2022imperceptible}, follows a more classical CNN-based design. Both are evaluated as embedding engines for our watermarking framework.
\subsubsection*{Stegformer}
This model follows a U-Net-inspired architecture \cite{ronneberger2015u}, where an autoencoder is used to encode a full image into a cover image and reconstruct it upon decoding. It includes a self-attention mechanism to enhance feature preservation and embedding quality (Figure \ref{fig:certification_examples}e-f).
\subsubsection*{SteGuz}
This model uses symmetry-aware CNNs to perform the embedding and recovery of a hidden image (Figure \ref{fig:certification_examples}c-d). The architecture includes a preprocessing block, an encoder for information embedding, and a decoder for marker recovery. 
To ensure that the embedding process maintains the visual fidelity of the cover image, SteGuz introduces a custom loss function based on two image similarity metrics:
\begin{itemize}[leftmargin=*]
\item \textbf{PSNR} (Peak Signal-to-Noise Ratio), which quantifies the ratio between the power of a signal and the power of corrupting noise, reflecting the fidelity between two compared images \cite{huynh2008scope}. It is defined as:
\begin{equation}
\small
    \text{PSNR}(x, y) = 10 \cdot \log_{10} \left( \frac{MAX_x^2}{\text{MSE}(x, y)} \right)
\end{equation}
where $x$ and $y$ are the original and reconstructed images respectively, both of size $m \times n$; $MAX_x$ is the maximum possible pixel value of the image (255 for 8-bit grayscale images); and MSE is the Mean Squared Error between $x$ and $y$, defined as:
\begin{equation}
\small
    \text{MSE}(x, y) = \frac{1}{mn} \sum_{i=1}^{m} \sum_{j=1}^{n} \left[ x(i,j) - y(i,j) \right]^2
\end{equation}
In this formulation, $x(i,j)$ and $y(i,j)$ denote the pixel intensities at position $(i,j)$ in the images $x$ and $y$, respectively. The MSE measures the average squared difference between corresponding pixels, and PSNR expresses the result in decibel scale \cite{avcibas2002statistical}.

\item \textbf{SSIM} (Structural Similarity Index Measure), which evaluates perceptual similarity between two images by comparing local patterns of pixel intensities normalized for luminance and contrast \cite{wang2004image}. It is computed as:
\begin{equation}
\small
    \text{SSIM}(x, y) = \frac{(2\mu_x \mu_y + C_1)(2\sigma_{xy} + C_2)}{(\mu_x^2 + \mu_y^2 + C_1)(\sigma_x^2 + \sigma_y^2 + C_2)}
\end{equation}
where $\mu_x$, $\mu_y$ are the local means, $\sigma_x^2$, $\sigma_y^2$ are the variances, and $\sigma_{xy}$ is the covariance between the two images. Constants $C_1$ and $C_2$ are used to stabilize the division in case of weak denominators.
\end{itemize}

\begin{table*}[t]
\centering
\caption{Summary of applied image manipulations and exact parameter values.}
\label{tab:manipulations}
\begin{tabular}{|l|l|l|l|c|}
\hline
\textbf{Manipulation Class} & \textbf{Manipulation Type} & \textbf{Parameter} & \textbf{Values} & \textbf{\# Samples} \\
\hline
\multirow{2}{*}{Compression} 
    & JPEG  & Quality Factor ($Q_F$) & 100, 99, 90, 80 & 3308 \\
    & WebP  & Quality Factor ($Q_F$) & 100, 99, 90, 80 & 3308 \\
\hline
Resizing 
    & Resizing & Scaling Factor ($R_F$) & \makecell[l]{99.9\%, 97.5\%, 95\%, 90\%,\\85\%, 75\%, 65\%, 50\%} & 6616 \\
\hline
Gaussian Noise Addition  
    & Gaussian Noise & Standard Deviation ($\sigma_G$) & \makecell[l]{2, 4, 6, 8, 10,\\16, 25, 32} & 6616 \\
\hline
Salt \& Pepper Noise Addition 
    & Salt \& Pepper Noise & Corruption Probability ($P_{SP}$) & \makecell[l]{(0.01, 0.3), (0.03, 0.1), (0.1, 0.03), (0.3, 0.01),\\(0.01, 0.01), (0.03, 0.03), (0.1, 0.1), (0.3, 0.3)} & 6616 \\
\hline
\multirow{2}{*}{Blurring} 
    & Gaussian Blur & Kernel Size ($K_G$) & 3, 5, 7, 9 & 3308 \\
    & Median Blur   & Kernel Size ($K_M$) & 3, 5, 7, 9 & 3308 \\
\hline
Sharpening 
    & Sharpening & Intensity Factor ($S_F$) & \makecell[l]{0, 0.001, 0.01, 0.05,\\0.1, 0.5, 0.75, 1} & 6616 \\
\hline
Morphing 
    & FaceMorpher & Blending Factor ($\alpha_M$) & 0.9 & 6616 \\
\hline
\end{tabular}
\end{table*}

\subsection{Image Manipulations}

To evaluate the resilience of the proposed integrity verification mechanism, we apply a series of controlled manipulations to the certified stego-image, simulating real-world modifications that may occur after issuance. These perturbations include both unintentional degradations (e.g., compression, resizing) and deliberate alterations (e.g., morphing, noise injection), allowing us to assess their impact on the embedded integrity marker. The exact parameters employed for each manipulation are reported in Table \ref{tab:manipulations}.

\subsubsection*{Compression} Images may undergo re-encoding during digital storage or transmission, leading to quality degradation. Additionally, compression artifacts often emerge when images are processed through automated verification systems, where scanned photos are stored or analyzed in varying formats. To examine this effect, we apply JPEG and WebP compression, controlled by the quality factor ($Q_F$), where $Q_F \in [80, 100]$. Lower values introduce moderate compression artifacts, while higher values result in minimal to no compression loss.
\subsubsection*{Resizing} ICAO-compliant images may be rescaled for different document formats, online submissions, or storage. To evaluate the resizing impact, each image is downscaled according to a resizing factor ($R_F$), where $R_F \in [50\%, 99.9\%]$. Higher values result in minor resizing effects; lower values introduce severe downscaling, causing loss of detail. The image is then restored to its original dimension to observe potential degradation in the embedded marker.

\subsubsection*{Noise Addition} Low-bitrate encoding, repeated compression cycles, and scanning artifacts can introduce unwanted noise, affecting the overall integrity of an image. We simulate these effects using:
\begin{itemize}
    \item \textit{Gaussian noise}, where the standard deviation ($\sigma_G$) is varied in the range $\sigma_G \in [2, 32]$. Low values introduce minor pixel intensity variations, while high values lead to strong noise artifacts affecting fine details.
    \item \textit{Salt-and-pepper noise}, parameterized by a corruption probability pair ($P_{SP}$), where $P_{SP} = (P_{\text{Salt}}, P_{\text{Pepper}})$.The first value represents the probability of a pixel turning white (Salt), and the second represents the probability of turning black (Pepper). Higher probabilities create more visible pixel corruption.
\end{itemize}

\subsubsection*{Blurring} Some post-processing techniques apply smoothing to remove noise or artifacts, which may also interfere with hidden data. We consider:
\begin{itemize}
    \item \textit{Gaussian blur}, which applies a weighted average of neighboring pixels, progressively diffusing fine details and potentially spreading steganographic patterns across a wider area. The kernel size ($K_G$) is selected from $K_G \in \{3, 5, 7, 9\}$, with larger values causing stronger blurring.
    \item \textit{Median blur}, which replaces each pixel with the median of its surrounding values, preserving edges better than Gaussian blur but disrupting embedded information by altering local pixel distributions. The kernel size ($K_M$) is chosen from $K_M \in \{3, 5, 7, 9\}$, where higher values increase the filtering effect.
\end{itemize}

\subsubsection*{Sharpening} Certain document processing tools enhance image clarity by artificially increasing edge contrast. The sharpening intensity factor ($S_F$) is adjusted within the range $S_F \in [0, 1]$. Low values produce no visible sharpening, while high values apply strong edge enhancement, which may introduce artificial artifacts.

\subsubsection*{Morphing} Unlike previous transformations, which may occur unintentionally, morphing is a deliberate biometric attack designed to deceive identity verification systems. In our study, we utilized \textit{FaceMorpher}\footnote{\url{https://github.com/alyssaq/face_morpher}}, an open-source tool based on facial landmarks to blend faces and create realistic morphed images. The blending factor ($\alpha_M$) controls the degree of fusion between two source images; we set $\alpha_M = 0.9$ in our experiments, favoring the stego-identity while subtly incorporating features of the second.
Therefore, while the outcome can be easily attributed to the most contributing individual, the second still has the chance to pass the identity verification \cite{korshunov2013using}. This choice maximizes the attack success rate for the first individual while maintaining plausible deniability in border control scenarios thanks to an increased realism compared, for instance, to $\alpha_M = 0.5$, which may produce morphs too distant from either original biometric template.

\subsection{Image Quality And Manipulation Assessment}

To assess the impact of image manipulations on the embedded integrity marker, we used three full-reference image quality metrics: Peak Signal-to-Noise Ratio (PSNR), Structural Similarity Index Measure (SSIM), and Mean Squared Error (MSE). The mathematical definitions of these metrics are provided in Section \ref{subsec:stega}.
Here, the focus is on their interpretation. PSNR and SSIM increase with image similarity, while MSE increases with distortion. An SSIM value close to 1 indicates high structural similarity; PSNR values above 50 dB typically correspond to minimal degradation. In contrast, higher MSE values reflect stronger pixel-wise differences.
These metrics allow us to quantify how much the hidden marker is degraded after manipulation, and to evaluate whether an image has been altered (Figures \ref{fig:steguz_manipulations} and \ref{fig:stegformer_manipulations}).

\subsection{Classification protocol}

To train a model capable of identifying the type of manipulation applied to the image from the revealed secret, we employed ResNet-50 \cite{koonce2021resnet}, pre-trained on ImageNet \cite{deng2009imagenet}, as a backbone for feature extraction. We set the classification problem to seven classes: compression, resize, blur, gaussian noise, salt and pepper noise, sharpening and morph generation.

To perform the classification from the extracted embeddings, we concatenated a sequence of fully connected layers. First, a linear layer reduces the dimensionality from $2048$ to $512$ units. Next, a ReLU activation function is introduced to add non-linearity. Then, a dropout layer with an activation probability of $0.5$ is inserted. Finally, a second linear layer reduces the output vector into the space of the target manipulation classes.

To assess the reliability of the classification model, we employed 70\% of the user identities in the dataset (i.e., \textit{578}) as the training set for fine-tuning and 30\% as the test set (i.e., \textit{249}). This allowed us to keep the manipulation classes balanced in the training and test sets and to simulate a real-world application context where the specific user face information is not known in training.

We evaluated the classification performance through metrics typically employed in pattern recognition: accuracy, precision, recall, and F1 Score.
To assess the ability to generalize across different embedding models and under previously unseen manipulation conditions, we designed four evaluation protocols:
\begin{itemize}[leftmargin=*]
    \item \textit{Intra-stega and intra-manipulation scenario:} both training and test images are embedded using the same model and subjected to the same types and strengths of manipulations.
    \item \textit{Cross-stega scenario:} training and test images are embedded using different steganography-based models, while the manipulation types remain consistent.
\end{itemize}

It is important to highlight that while in a practical certification scenario, the embedding method would typically remain fixed throughout the system lifecycle, the cross-stega analysis can be useful to assess the generalizability and robustness of the proposed manipulation detection approach. In fact, it simulates potential real-world inconsistencies, such as re-certification with a different method or interoperability between systems using distinct embedding techniques. In addition, it provides information on the transferability of learned features between embedding strategies, which is essential for scalable or future-proof implementations. 

Additionally, we applied two sub-protocols:
1) the first one, called P8-8, involves training and testing on the same eight variations of each manipulation type (e.g., levels of noise or compression); 2) the second, called P6-8, , involves training on six variations per manipulation type, while testing includes all eight—introducing two never-seen-before variations for each manipulation during testing.

\section{Results}\label{sec:results}

In this section, we present the results obtained from the previously described experiments. In Section \ref{subsec:stego}, we report the results obtained by analyzing the impact of the modifications in the images employed on the integrity verification process. In Section \ref{subsec:classifier}, we discuss the capabilities of the classification model in detecting unauthorized modifications and distinguishing between different types of alterations. 

\subsection{Image quality and manipulation assessment results}\label{subsec:stego}
To evaluate the feasibility of using secret image hiding and recovery for manipulation assessment, we analyzed the quality of the stego and recovered images before/after manipulations. 

As shown in Table \ref{tab:recovery_hiding_quality}, the average quality of the certified images and the revealed images, quantified by SSIM, MSE and PSNR, demonstrates that the distortion introduced by the certifying watermark is minimal and that the recovery processes produce high-fidelity outputs. This confirms that the embedding and extraction mechanisms, even when relying on different steganographic approaches (Steguz and Stegformer), do not introduce significant artifacts in both phases. For instance, the SSIM values remain above 0.92 for both models and stages, indicating that the certification and recovery processes do not noticeably corrupt the final images and are acceptable from a practical perspective. Comparing the two models, Stegformer provided better performance considering all the analyzed quality assessment metrics.

\begin{figure}[t]
    \centering
    \begin{subfigure}{0.09\textwidth}
        \includegraphics[width=\linewidth]{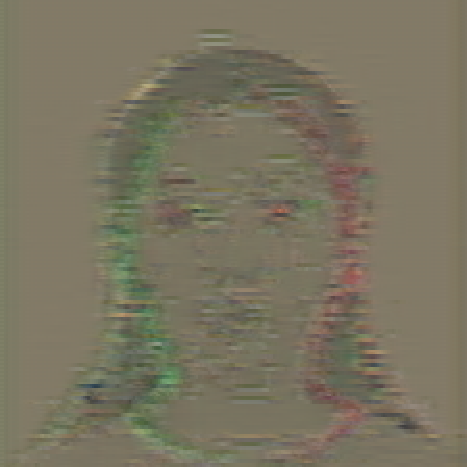}
        \caption{}
    \end{subfigure}
    \begin{subfigure}{0.09\textwidth}
        \includegraphics[width=\linewidth]{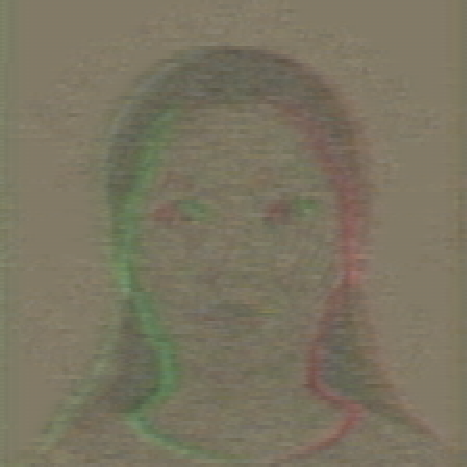}
        \caption{}
    \end{subfigure}
    \begin{subfigure}{0.09\textwidth}
        \includegraphics[width=\linewidth]{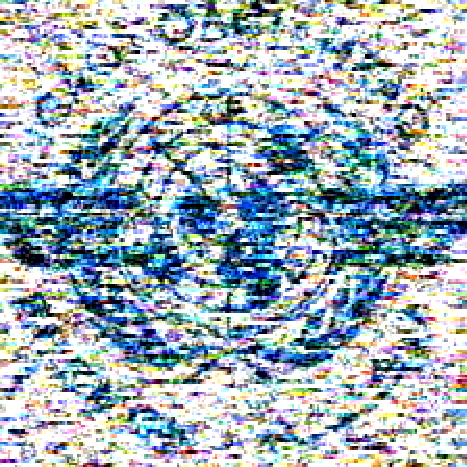}
        \caption{}
    \end{subfigure}
    \begin{subfigure}{0.09\textwidth}
        \includegraphics[width=\linewidth]{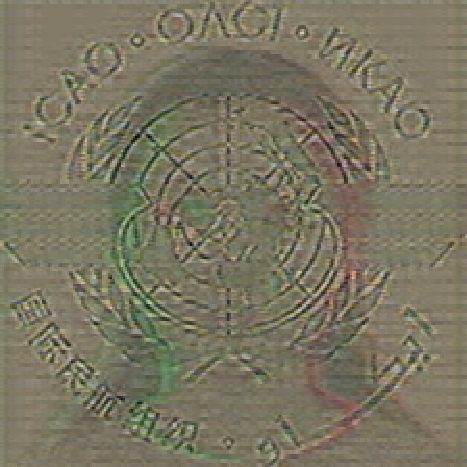}
        \caption{}
    \end{subfigure}\\
    \begin{subfigure}{0.09\textwidth}
        \includegraphics[width=\linewidth]{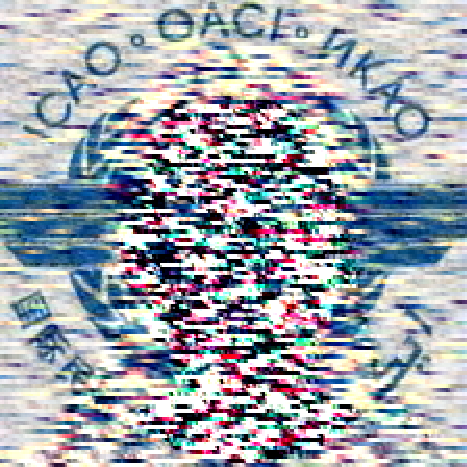}
        \caption{}
    \end{subfigure}
    \begin{subfigure}{0.09\textwidth}
        \includegraphics[width=\linewidth]{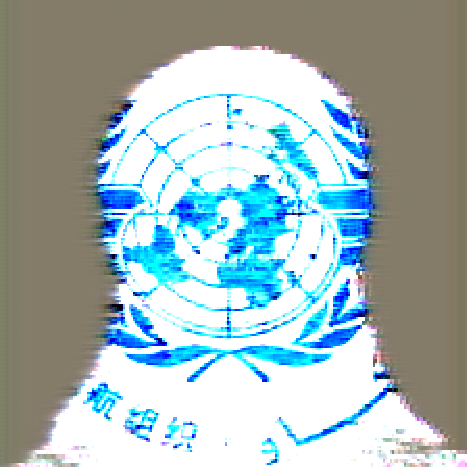}
        \caption{}
    \end{subfigure}
    \begin{subfigure}{0.09\textwidth}
        \includegraphics[width=\linewidth]{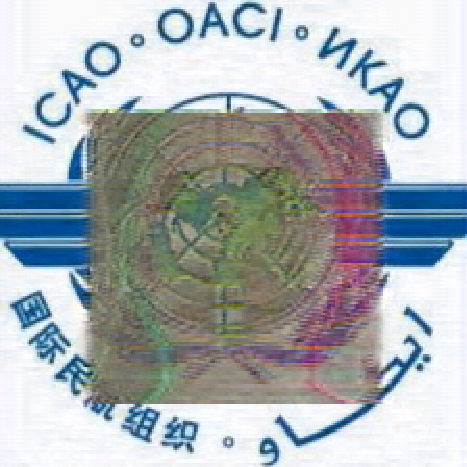}
        \caption{}
    \end{subfigure}
    \caption{Recovered secret images examples using Steguz after applying manipulations: a) JPEG compression ($Q_{F}=80$), b) Gaussian blur ($K_{G}=7$), c) Gaussian noise ($\sigma=8$), d) resize ($R_{F}=85\%$), e) salt \& paper noise ($P_{SP}=\left(0.3, 0.01\right)$), f) sharpening ($S_{F}=0.5$), g) morphing ($\alpha_{M}=0.9$).}
    \label{fig:steguz_manipulations}
\end{figure}

\begin{figure}[t]
    \centering
    \begin{subfigure}{0.09\textwidth}
        \includegraphics[width=\linewidth]{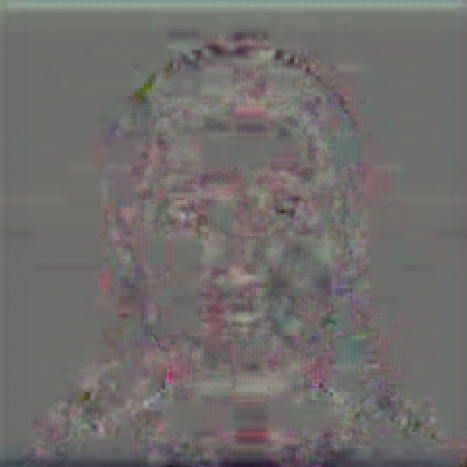}
        \caption{}
    \end{subfigure}
    \begin{subfigure}{0.09\textwidth}
        \includegraphics[width=\linewidth]{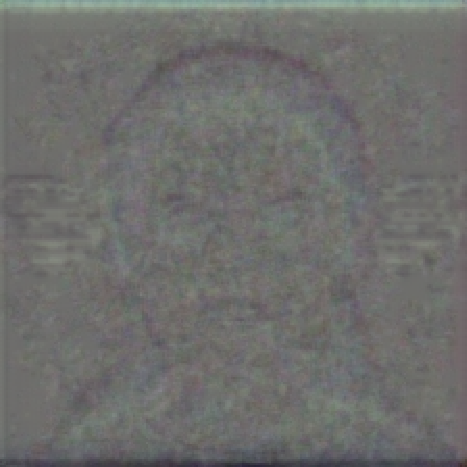}
        \caption{}
    \end{subfigure}
    \begin{subfigure}{0.09\textwidth}
        \includegraphics[width=\linewidth]{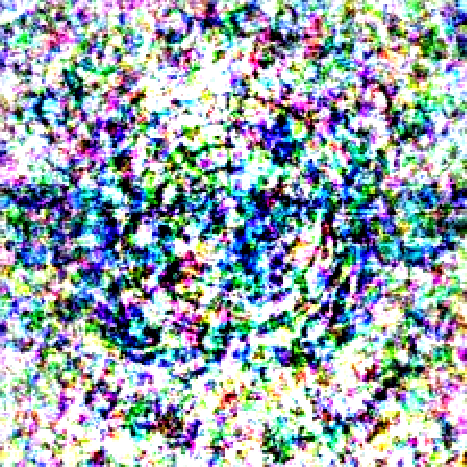}
        \caption{}
    \end{subfigure}
    \begin{subfigure}{0.09\textwidth}
        \includegraphics[width=\linewidth]{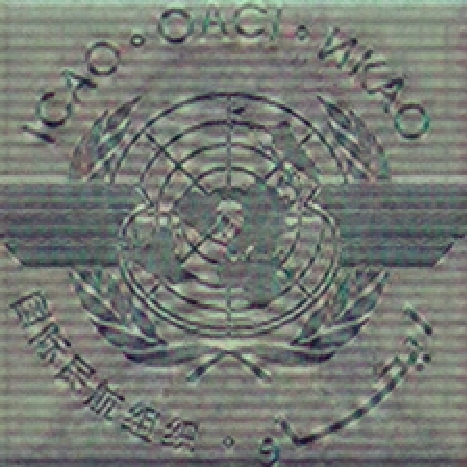}
        \caption{}
    \end{subfigure}\\
    \begin{subfigure}{0.09\textwidth}
        \includegraphics[width=\linewidth]{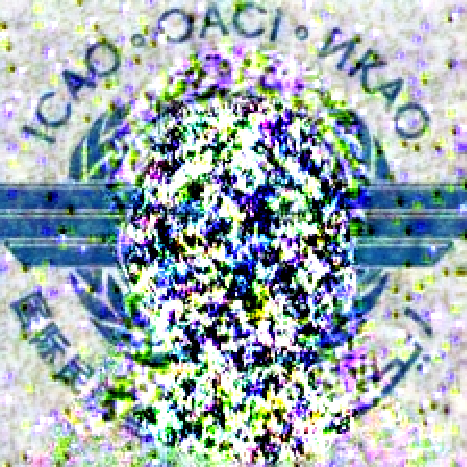}
        \caption{}
    \end{subfigure}
    \begin{subfigure}{0.09\textwidth}
        \includegraphics[width=\linewidth]{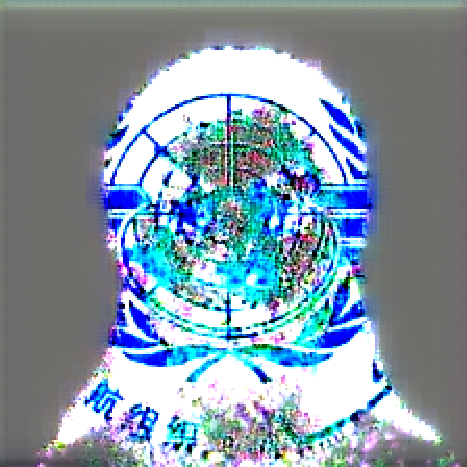}
        \caption{}
    \end{subfigure}
    \begin{subfigure}{0.09\textwidth}
        \includegraphics[width=\linewidth]{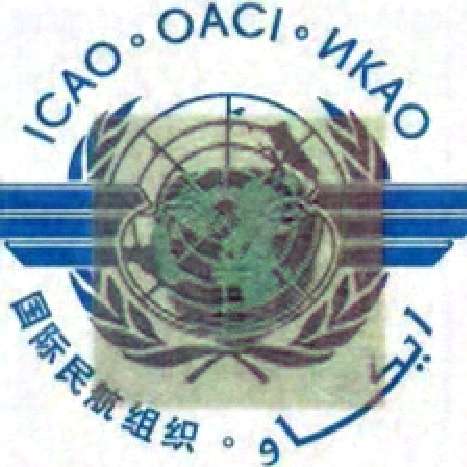}
        \caption{}
    \end{subfigure}
    \caption{Recovered secret images examples using Stegformer after applying manipulations: a) JPEG compression ($Q_{F}=80$), b) Gaussian blur ($K_{G}=7$), c) Gaussian noise ($\sigma=8$), d) resize ($R_{F}=85\%$), e) salt \& paper noise ($P_{SP}=\left(0.3, 0.01\right)$), f) sharpening ($S_{F}=0.5$), g) morphing ($\alpha_{M}=0.9$).}
    \label{fig:stegformer_manipulations}
\end{figure}


\begin{table}[t]
\centering
\caption{Difference between original image and certified image (task certifying) and between original and recovered secret image (task recovery) in terms of SSIM, MSE and PSNR. Values are mean ± standard deviation.}
\resizebox{\columnwidth}{!}{%
\begin{tabular}{|l|l|c|c|c|}
\hline
\textbf{Task} & \textbf{Model} & \textbf{SSIM} & \textbf{MSE} & \textbf{PSNR} \\
\hline
\multirow{2}{*}{Certifyng}   & Steguz     & 0.9266 ± 0.0097 & 123.27 ± 33.49 & 27.37 ± 1.14 \\
                          & Stegformer & 0.9696 ± 0.0035 & 7.23 ± 0.65 & 39.56 ± 0.40 \\
\hline                       
\multirow{2}{*}{Recovery} & Steguz     & 0.9389 ± 0.0015 & 97.91 ± 3.51 & 28.23 ± 0.15 \\
                          & Stegformer & 0.9508 ± 0.0013 & 35.35 ± 1.81 & 32.65 ± 0.23 \\

\hline
\end{tabular}%
}
\label{tab:recovery_hiding_quality}
\end{table}

\begin{figure}
    \centering
    \includegraphics[width=0.9\columnwidth]{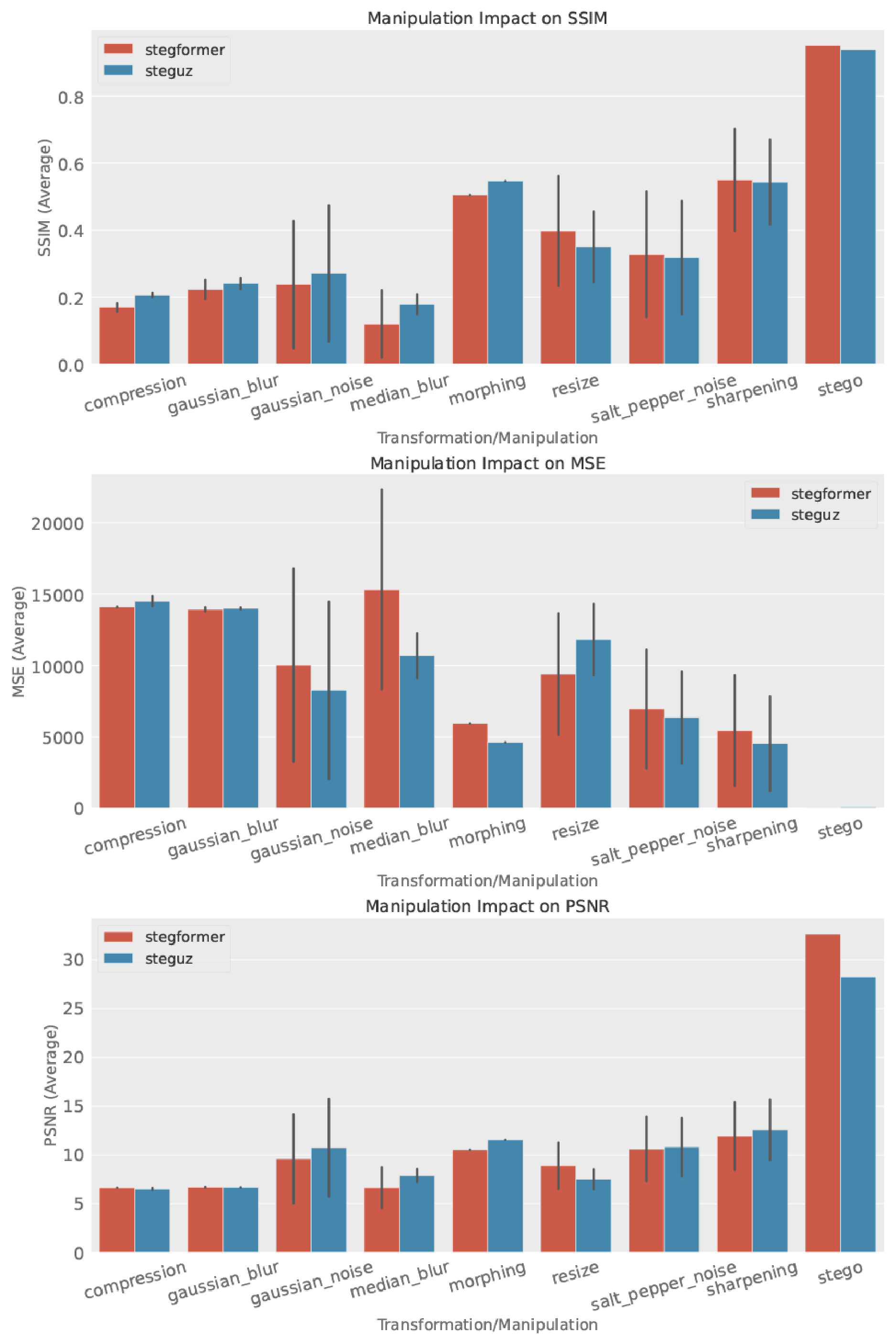}
    \caption{Effect of transformations on image recovery. Error bars reflect variation within each manipulation type.}
    \label{fig:results_general_effects_revelead}
\end{figure}

The impact of the manipulations on the secret image is, instead, appreciable and is shown in Figure \ref{fig:results_general_effects_revelead}.
In all three metrics (SSIM, MSE, PSNR), manipulations such as Gaussian noise, morphing, and salt-and-pepper noise result in a substantial drop in quality. For example, the SSIM for morphing and salt-and-pepper noise drops below 0.5 in several cases, with a corresponding spike in MSE. These quality degradations are not only measurable, but also visually perceptible (see Figures \ref{fig:steguz_manipulations} and \ref{fig:stegformer_manipulations}), confirming that the manipulations leave strong and consistent fingerprints on the recovered secret image.

This behavior is the foundation for the proposed manipulation classifier. Despite being built on different design principles (Stegformer aiming for generalization, and Steguz for robustness) both models react similarly to post-embedding manipulations, suggesting that the introduced artifacts are sufficiently distinctive to be exploited for classification.

To support interpretation in practical deployments, we define operational thresholds derived from the observed values in the unaltered (certified) case. Specifically, images with SSIM below 0.75 or PSNR below 22 dB are considered potentially manipulated.

\subsection{Classification Results}\label{subsec:classifier}

\begin{table}[t]
\centering
\caption{Classification performance in intra-stega and cross-stega scenarios (P8-8 sub-protocol).}
\resizebox{\columnwidth}{!}{
\begin{tabular}{|l|l|c|c|c|c|}
\hline
\textbf{Training set} & \textbf{Test Set} & \textbf{Accuracy} & \textbf{Precision} & \textbf{Recall} & \textbf{F1 Score} \\
\hline
\multirow{2}{*}{Stegformer} & Stegformer & 99.96\% & 99.96\% & 99.96\% & 99.96\% \\
                            & Steguz     & 71.92\% & 77.97\% & 71.92\% & 68.99\% \\
\hline
\multirow{2}{*}{Steguz}     & Steguz     & 99.95\% & 99.95\% & 99.95\% & 99.95\% \\
                            & Stegformer & 80.51\% & 86.16\% & 80.51\% & 77.35\% \\
\hline
\end{tabular}}
\label{tab:stego_comparison}
\end{table}

The classifier results reported in Table \ref{tab:stego_comparison} show that manipulations can be detected by analyzing the recovered secret image. The intra-stega results for protocol P8-8 demonstrate that most manipulated samples ($\ge$99.95\%) are correctly classified. However, the effects of manipulations vary depending on the embedding method used. In fact, the cross-stega results show an average performance drop of about 25\%.

\begin{figure*}[ht!]
    \centering

    \begin{subfigure}[b]{0.44\linewidth}
        \centering
        \includegraphics[width=0.85\textwidth]{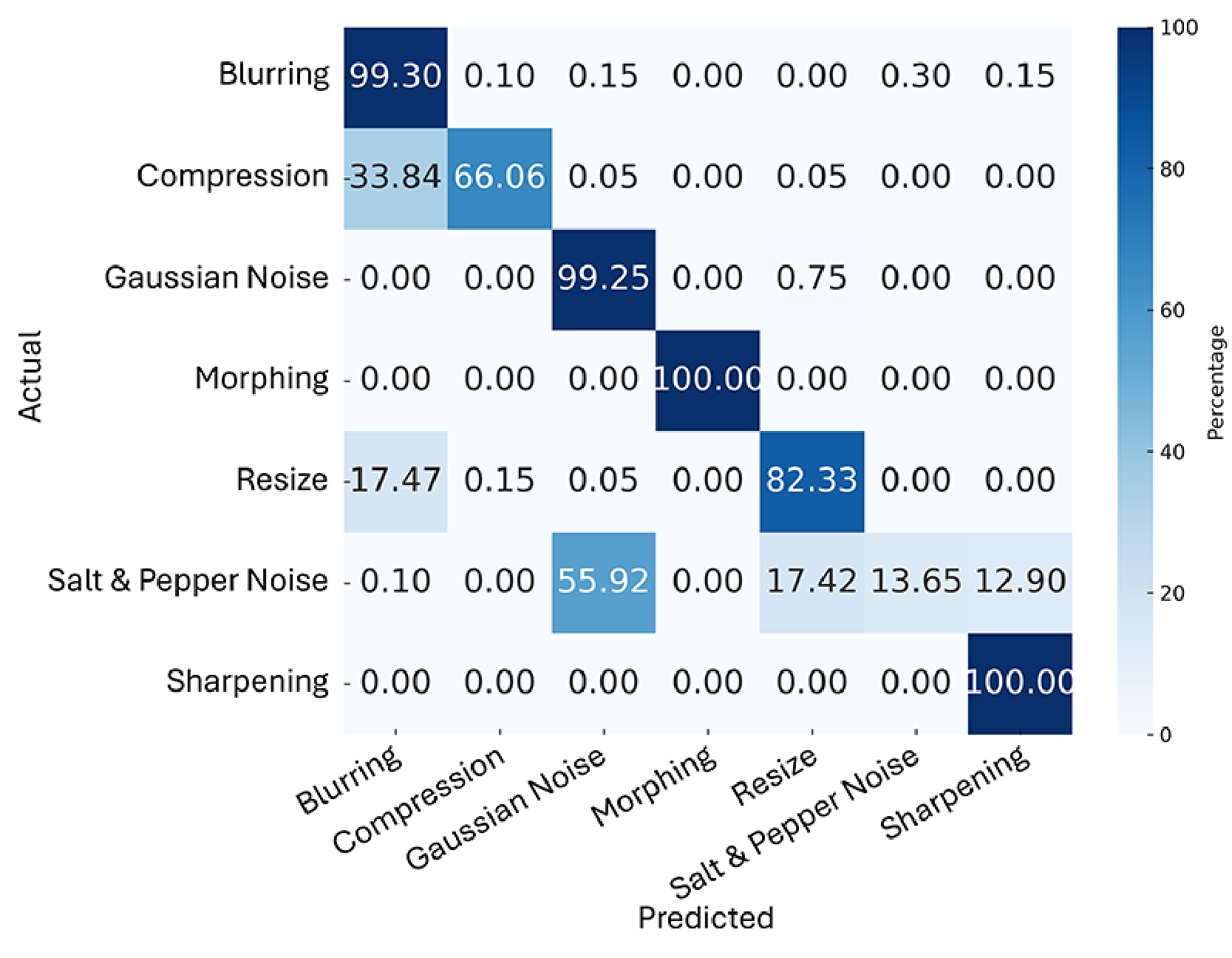}
        \caption{P6-8 - Trained on Steguz, tested on Stegformer}
        \label{fig:p68_steguz_cross}
    \end{subfigure}
    \hfill
    \begin{subfigure}[b]{0.44\linewidth}
        \centering
        \includegraphics[width=0.85\textwidth]{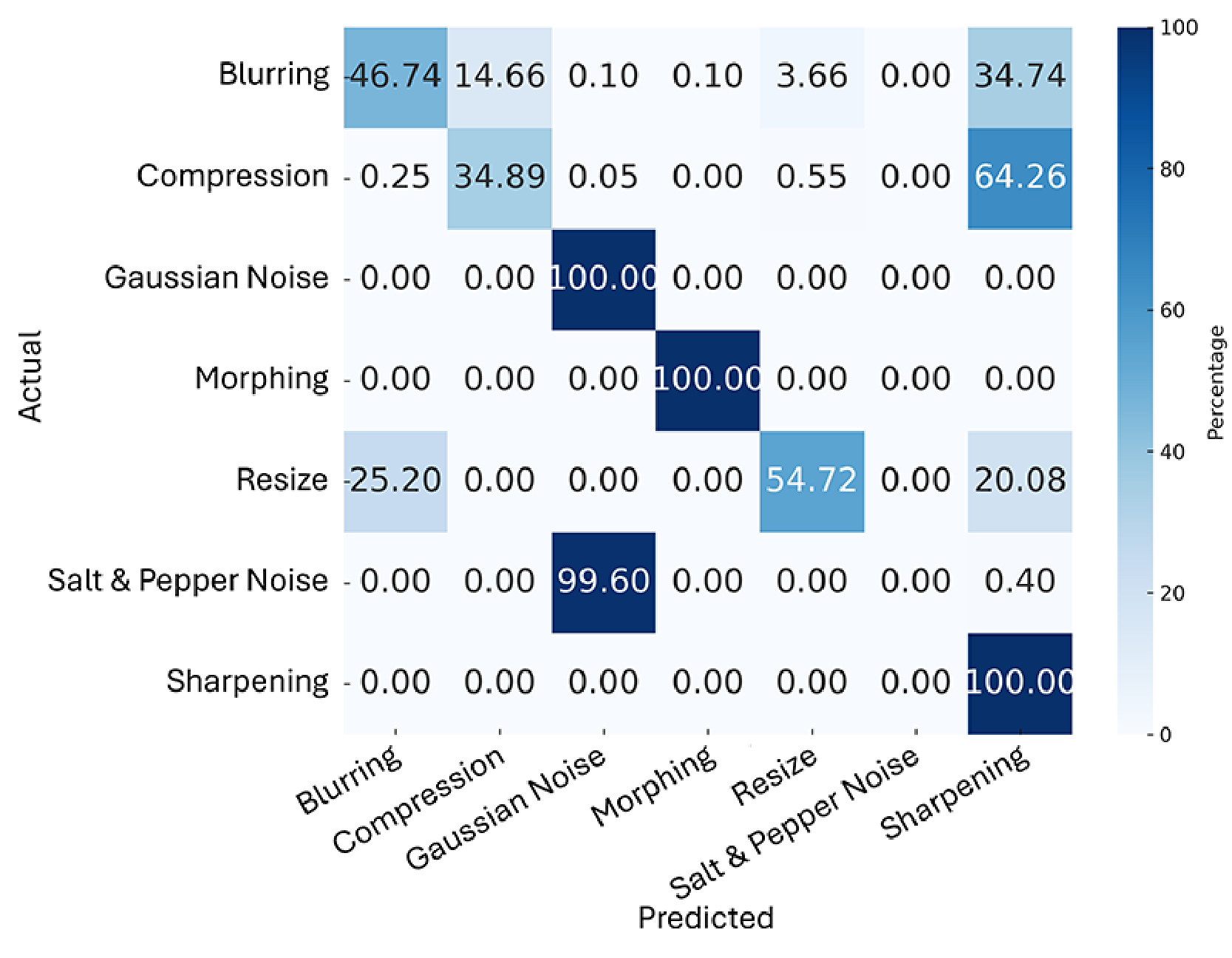}
        \caption{P6-8 - Trained on Stegformer, tested on Steguz}
        \label{fig:p68_stegformer_cross}
    \end{subfigure}

    \vspace{0.2cm}

    \begin{subfigure}[b]{0.44\linewidth}
        \centering
        \includegraphics[width=0.85\textwidth]{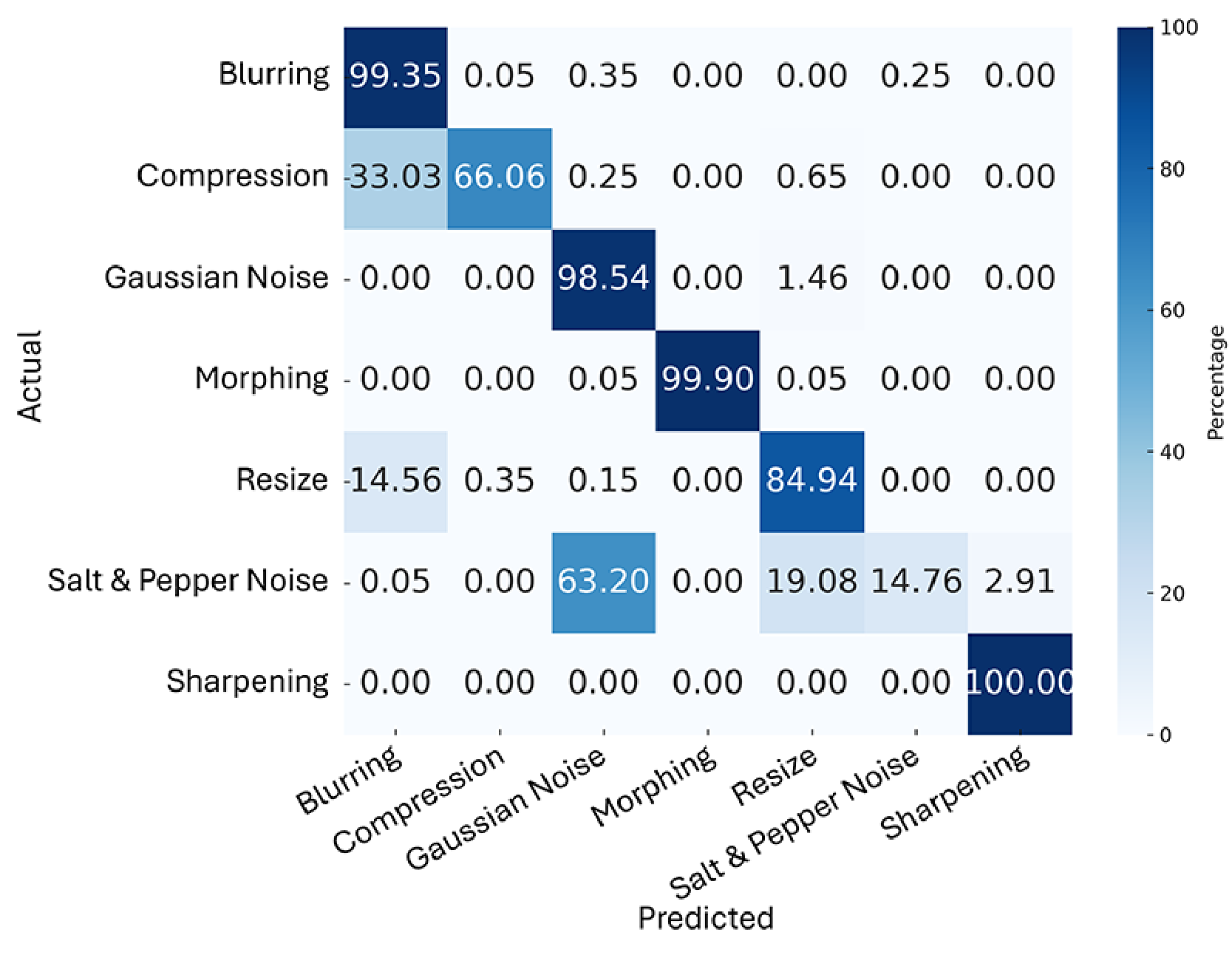}
        \caption{P8-8 - Trained on Steguz, tested on Stegformer}
        \label{fig:p88_steguz_cross}
    \end{subfigure}
    \hfill
    \begin{subfigure}[b]{0.44\linewidth}
        \centering
        \includegraphics[width=0.85\textwidth]{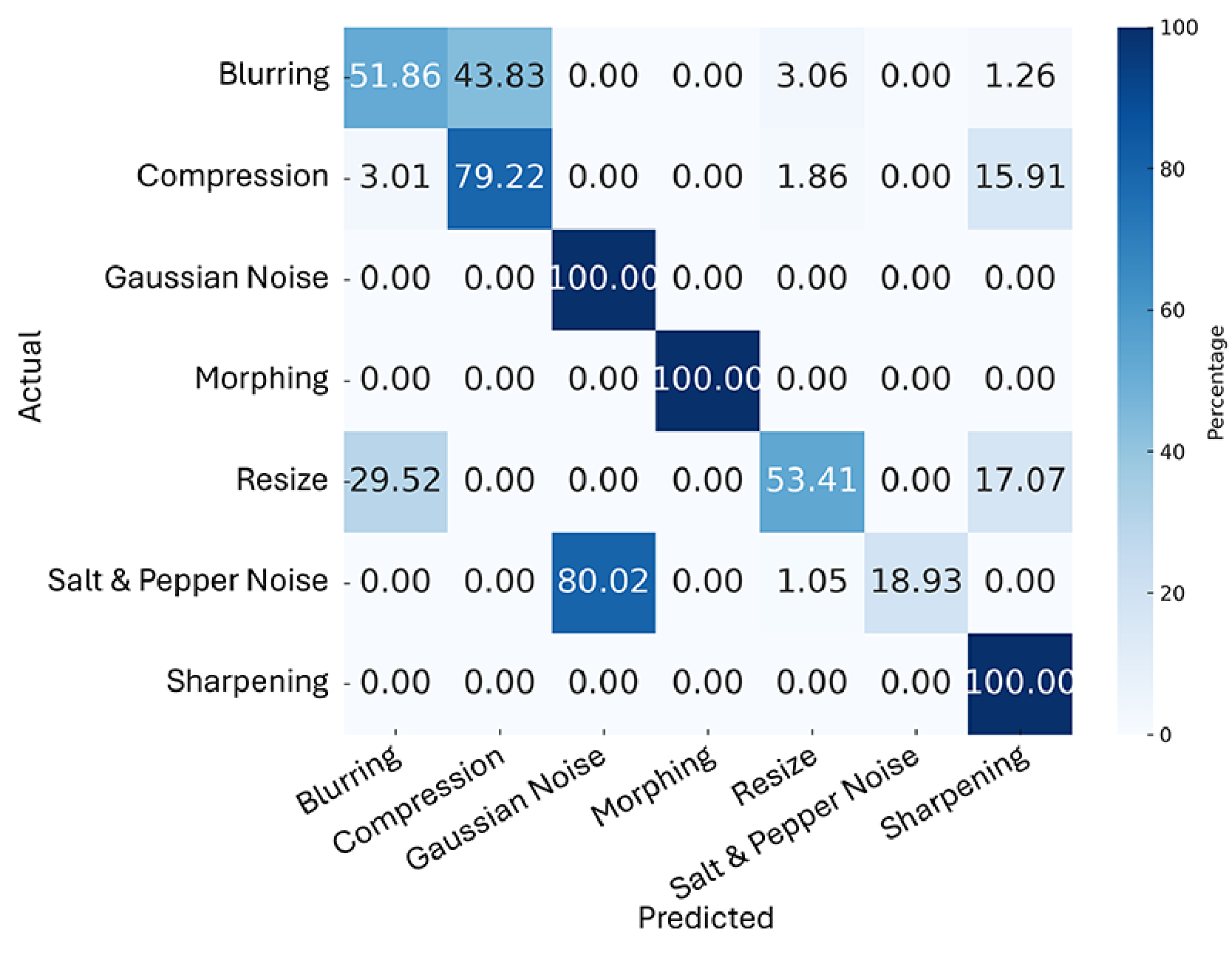}
        \caption{P8-8 - Trained on Stegformer, tested on Steguz}
        \label{fig:p88_stegformer_cross}
    \end{subfigure}

    \caption{Confusion matrices for Stegformer and Steguz models under cross-stega evaluation for  P6-8 and P8-8.}
    \label{fig:conf_matrices_cross}
\end{figure*}

A more detailed analysis through both the P8-8 and P6-8 sub-protocols highlights that the degradation in the generalization capability is strongly influenced by the type of manipulation applied, as shown by confusion matrices in Figure \ref{fig:conf_matrices_cross}. In particular, the classification is still reliable on manipulation classes that generate well-defined and consistent structural and chromatic visual patterns, such as morphing, sharpening, and the addition of Gaussian noise. Most classification errors in cross-stega scenario involve the salt \& pepper noise class being confused with Gaussian noise. Since these two types of manipulation are quite similar, such errors are potentially negligible in this application context. 

Other manipulations often misclassified are resize and compression. Specifically, samples altered through resizing are, in some cases, misclassified as blurring. Regarding compression, the recovered images tend to be misclassified as either blurring or sharpening, depending on the method used. 

In general, when comparing cross-stega P6-8 and P8-8 protocols, we observe that the performance drop primarily affects the model trained on images generated using Stegformer. In contrast, the model trained on Steguz maintains similar performance across both protocols. The latter is, therefore, able to generalize better on unknown variations.

In summary, the experimental results support the feasibility of using fragile watermarking, implemented via deep steganographic embedding, for image certification purposes. The proposed classification model was able to reliably identify the type of manipulation applied to the host image, even in challenging scenarios such as cross-stega settings and in the presence of unseen transformation variations. Despite the architectural and functional differences between the embedding models, the visual degradations produced on the revealed integrity marker were distinctive enough to enable generalization. This highlights the approach's potential for integration into real-world integrity verification pipelines, where robustness, interpretability, and scalability are critical requirements.

\section{Conclusions}\label{sec:conclusions}

This work proposed a fragile watermarking framework for integrity verification of ICAO-compliant biometric images, based on deep steganographic embedding. A known visual marker is embedded into the facial image and later recovered to detect possible post-issuance manipulations through visible degradation.
We evaluated this approach in the context of ICAO-compliant identity images, using two state-of-the-art steganography-based embedding models. A range of transformations, including compression, resizing, noise, and morphing, were applied to test the sensitivity and diagnostic value of the revealed marker. Beyond tamper detection, we assessed the feasibility of classifying the type of manipulation by analyzing patterns of degradation in the recovered image.
The findings demonstrate that steganography-based fragile watermarking can provide not only binary integrity verification but also actionable forensic information. To our knowledge, this is the first study to assess the use of standard deep steganographic models for this purpose in the context of biometric documents. To our knowledge, this is the first study to assess the use of standard deep steganographic models for this purpose in the context of document integrity.
Future work will explore additional embedding architectures, extend the method to other biometric modalities, and evaluate robustness under adversarial conditions.

\section*{Acknowledgment}

This work was partially supported by Project SERICS (PE00000014) under the NRRP MUR program funded by the EU - NGEU. Davide Ghiani's PhD grant is partly funded by Dedem SpA under the PNRR program.


\end{document}